\documentclass[runningheads]{llncs}

\usepackage{eccv}

\usepackage{eccvabbrv}

\usepackage{graphicx}
\usepackage{booktabs}
\usepackage{multirow}
\usepackage[accsupp]{axessibility}  

\usepackage[pagebackref,breaklinks,colorlinks,citecolor=eccvblue]{hyperref}

\usepackage{orcidlink}
\usepackage{tabularx}
\usepackage{array}
\newcolumntype{C}[1]{>{\centering\arraybackslash}p{#1}}

\usepackage{amsmath}

\usepackage{hyperref}
\begin{document}


\title{Learning Trimaps via Clicks for Image Matting} 

\titlerunning{Click2Trimap}

\author{Chenyi Zhang\inst{1} \and
Yihan Hu\inst{1}\and
Henghui Ding\inst{2}\and
\\
Humphrey Shi\inst{3}\and
Yao Zhao\inst{1}\and
Yunchao Wei\inst{1}\thanks{Corresponding author}}

\institute{Institute of Information Science, Beijing Jiaotong University \and
Nanyang Technological University \and
Georgia Tech $\&$ Picsart AI Research (PAIR)
\\
 {\tt\small chenyi22@bjtu.edu.cn}
}

\maketitle

\begin{abstract}
  Despite significant advancements in image matting, existing models heavily depend on manually-drawn trimaps for accurate results in natural image scenarios. However, the process of obtaining trimaps is time-consuming, lacking user-friendliness and device compatibility. This reliance greatly limits the practical application of all trimap-based matting methods. To address this issue, we introduce \textbf{Click2Trimap}, an interactive model capable of predicting high-quality trimaps and alpha mattes with minimal user click inputs. Through analyzing real users' behavioral logic and characteristics of trimaps, we successfully propose a powerful iterative three-class training strategy and a dedicated simulation function, making Click2Trimap exhibit versatility across various scenarios. Quantitative and qualitative assessments on synthetic and real-world matting datasets demonstrate Click2Trimap's superior performance compared to all existing trimap-free matting methods. Especially, in the user study, Click2Trimap achieves high-quality trimap and matting predictions in just an average of 5 seconds per image, demonstrating its substantial practical value in real-world applications. You are welcome to try our demo at \href{https://github.com/ChenyiZhang007/Click2Trimap}{\url{https://github.com/ChenyiZhang007/Click2Trimap}}.
\end{abstract}
  
\section{Introduction}
\label{sec:intro}

\begin{figure}[!t]
  \centering
  \begin{minipage}{\textwidth}
  \centering
    \includegraphics[width=0.95\textwidth]{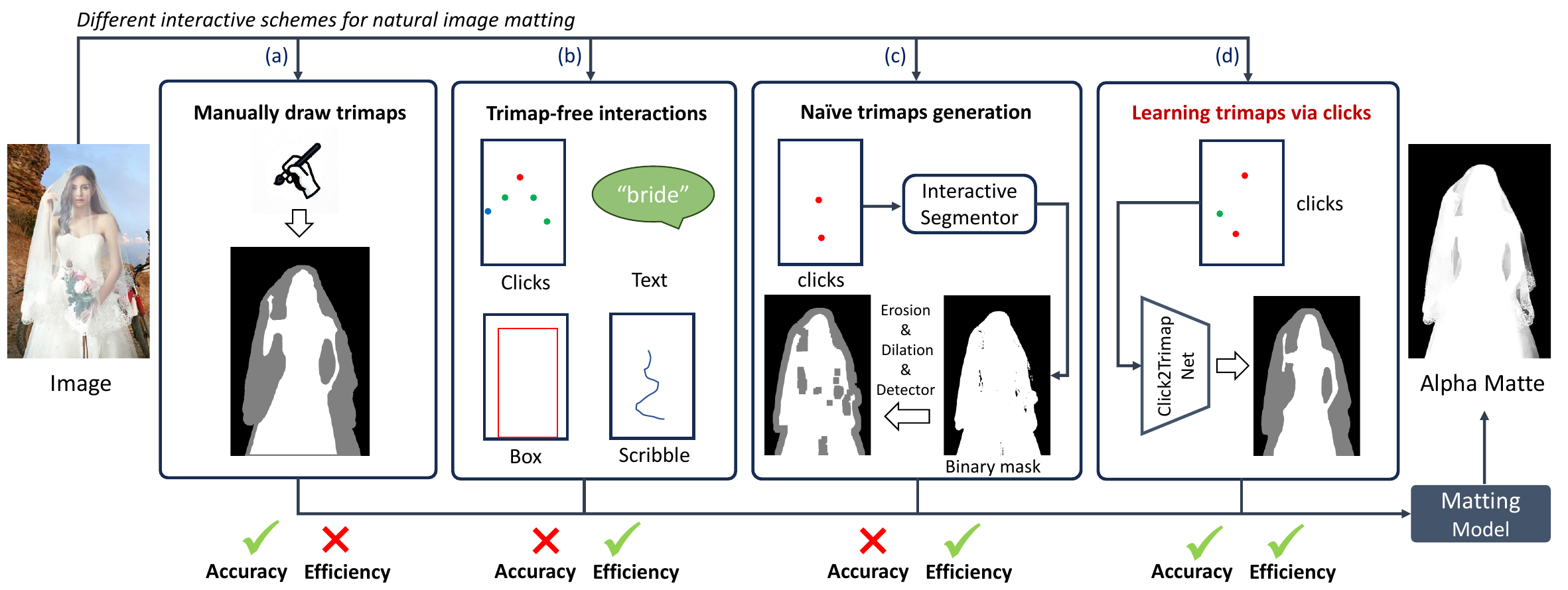}
    \caption{Illustration of different interactive schemes for natural image matting. \protect\begin{tikzpicture}
       \protect\node[inner sep=0pt, text=white] (charA) {A};
        \protect\fill[red] (charA.center) circle (0.05);
    \protect\end{tikzpicture}, \protect\begin{tikzpicture}
       \protect\node[inner sep=0pt, text=white] (charA) {A};
        \protect\fill[green] (charA.center) circle (0.05);
    \protect\end{tikzpicture}, and \protect\begin{tikzpicture}
       \protect\node[inner sep=0pt, text=white] (charA) {A};
        \protect\fill[blue] (charA.center) circle (0.05);
    \protect\end{tikzpicture} refer to \textit{foreground}, \textit{unknown}, and \textit{background} clicks, respectively. Compared with previous solutions, we proposed a new interactive scheme, i.e., learning trimaps via clicks,  which simultaneously achieves both accuracy and efficiency.}
    \label{fig:task}
  \end{minipage}

  \vspace{5mm} 

  \begin{minipage}{\textwidth}
    \centering
    \includegraphics[width=\textwidth]{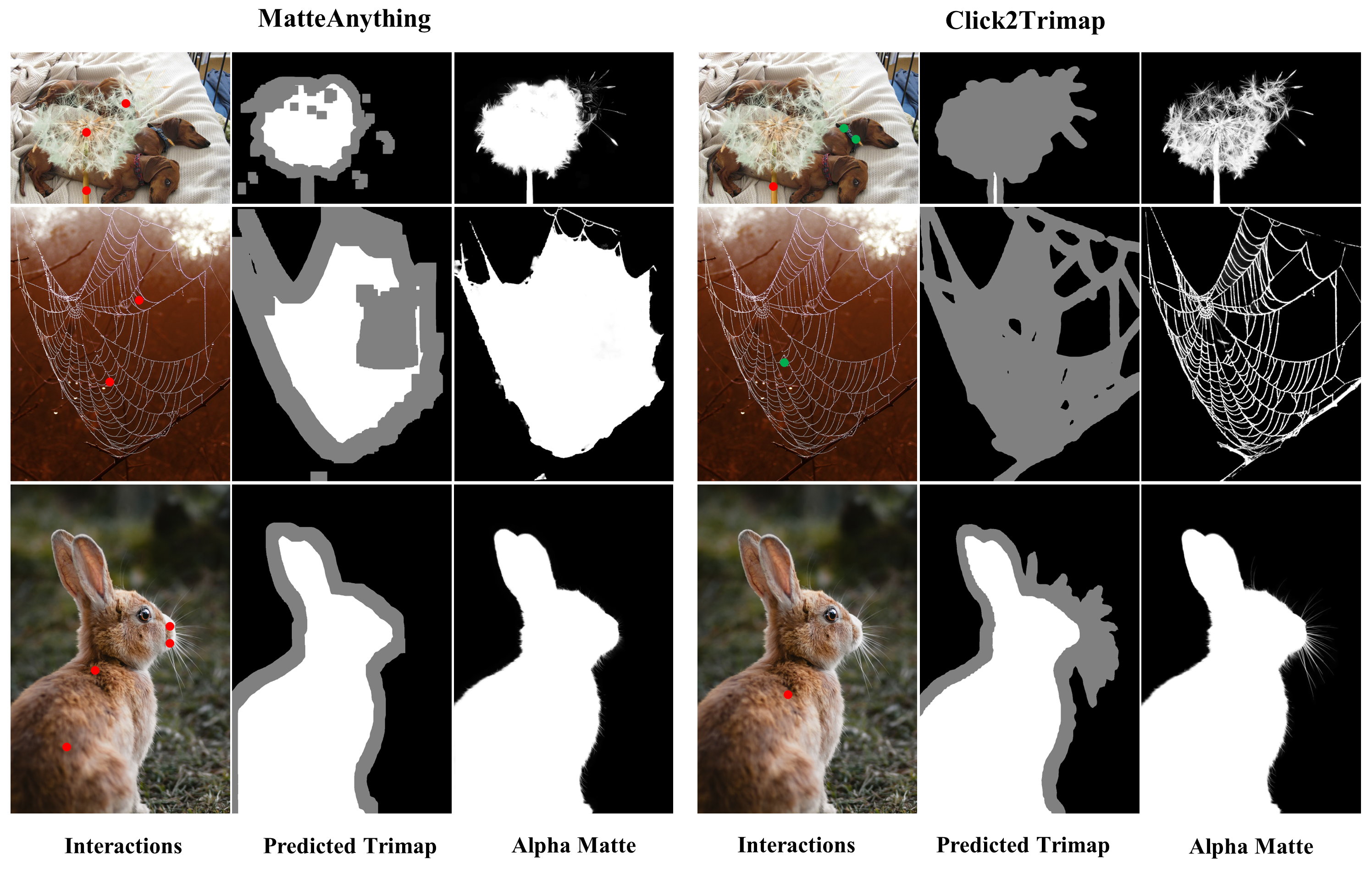}
    \caption{Qualitative comparison between our method and the state of the art method, \ie, MatteAnything\cite{yao2023matte}, on challenging cases.}
    \label{fig:fig2}
  \end{minipage}
  \vspace{-2em}
\end{figure}

Natural image matting, a fundamental task in the field of computer vision, holds substantial significance with broad applications in real-world scenarios such as image editing, video conferencing, and the movie industry. The goal of matting is to predict the transparency of target objects. Given an input image $I$, the matting problem considers it as the composite of a foreground layer $F$ and a background layer $B$, and the value at each pixel $i$ is defined as a convex combination of $F_i$ and $B_i$ by using the alpha matte $\alpha_i\in[0,1]$. Thus, $I_i$ can be expressed by :
\begin{equation}
    I_i = {\alpha}_i F_i + (1-\alpha_i)B_i. 
\end{equation}
If $\alpha_i =1$ or $0$, the pixel $I_i$ is a pure foreground pixel or a pure background pixel, respectively. 

Given the values of $\alpha$, foreground $F$, and background $B$ are all unknown, image matting is a highly ill-posed problem. In addressing this complexity, the prevailing natural image matting methods require user guidance as crucial clues to facilitate the procedure. A common approach to provide this user guidance involves the utilization of the trimap~\cite{xu2016deep}, an image labeling each pixel as \textit{foreground}, \textit{background}, or \textit{unknown}, as described in \cref{fig:task} (a). This approach has been integrated into several matting methods and achieved remarkable matting performance \cite{xu2017deep, li2020natural, park2022matteformer, yao2023vitmatte, liu2023rethinking, fbaMatting20}. However, the preparation of such dense trimaps is laborious and resource-intensive. For example, the average time investment for drawing a trimap exceeds 200 seconds, with more intricate cases surpassing 10 minutes~\cite{wei2021improved, yang2022unified}, which significantly constrains the practical utility and research scope of trimap-based matting methods. Consequently, various methods \cite{yang2020smart,wei2021improved,ding2022deep} have been proposed to realize natural image matting with a more manageable labor cost by leveraging more sparse forms of guidance, such as clicks, boxes, text, and scribbles, as depicted in \cref{fig:task} (b). However, those sparse guidances often lack sufficient guidance information, leading to a noticeable accuracy gap when compared to trimap-based methods. More recently, an alternative approach to providing guidance has been introduced by MatteAnything \cite{yao2023matte}, as decribed in \cref{fig:task} (c). Specifically, they first use a powerful interactive segmentation model, SAM~\cite{Kirillov_2023_ICCV}, to obtain an initial mask with minimal user clicks, then a post-processing that includes erosion, dilation and a multi-modality detector \cite{liu2023grounding} converts this initial mask into a trimap. Subsequently, the generated trimap is fed into a trimap-based matting model to predict the final alpha matte. This approach effectively combines the strengths of trimap-free and trimap-based methods, offering a balance between efficiency and accuracy. However, despite its advancements compared to trimap-free methods, MatteAnything still lags behind trimap-based methods in terms of precision. This limited accuracy can be attributed to significant issues in the quality of the generated trimaps, as demonstrated in \cref{fig:fig2}. Therefore, it remains to explore a method that can convert sparse user inputs into high-quality trimaps, further bridging the gap between efficiency and accuracy.

In this paper, we introduce \textbf{Click2Trimap}, a novel model specifically designed for efficient and user-friendly natural image matting based on sparse clicks. Our goal is to generate high-quality trimaps using minimal user inputs, enabling seamless integration with trimap-based matting methods, and enhancing the accuracy of alpha mattes, as shown in \cref{fig:task} (d). Achieving precise trimap predictions poses a primary challenge due to the ambiguous nature of unknown regions, which often exhibit overlap with both foreground and background regions. To generate accurate trimaps, a comprehensive understanding of the user clicks is necessary. Therefore, we formulate the task of trimap prediction with user clicks as a three-class interactive segmentation task. Considering the widespread use and proven effectiveness \cite{chen2022focalclick, liu2023simpleclick, chen2021conditional, sofiiuk2022reviving} of Iterative Training Strategy (ITS) \cite{mahadevan2018iteratively} in the field of binary interactive segmentation, we specially design an Iterative Three-class Training Strategy (ITTS) for our task. By parsing a trimap into three binary mask and analyzing them separately, ITTS facilitates the iterative training of a three-class interactive segmentation model.

In addition, the accuracies of different classes in the trimap have distinct effects on the subsequent prediction of the alpha matte. For instance, falsely predicting pixels as either the \textit{foreground} or \textit{background} classes can lead to incorrect guidance for the alpha matte prediction, while misclassifying pixels as the \textit{unknown} class has a comparatively lower impact on the alpha matte prediction. Therefore, ensuring a high recall rate for the \textit{unknown} class is essential for accurate alpha matte prediction. Considering this, we introduce a strategy named Conditioned Unknown Prioritized Simulation (CUPS), prioritizing the simulation of clicks for the \textit{unknown} class to meet the distinctive requirements of the alpha matte prediction.

Comprehensive experiments on various popular synthetic and real-world matting datasets demonstrate Click2Trimap's superiority over existing click-based matting methods, both quantitatively and qualitatively. We further conduct a user study to validate Click2Trimap's remarkable efficiency in interactions, substantially reducing the time needed to obtain precise alpha mattes. Click2Trimap achieves accuracy comparable to trimap-based matting methods while preserving the cost-effectiveness of trimap-free matting methods. \textbf{Notably}, with the trimap generated by our method, the performance of alpha matte prediction is comparable to that of manually annotated trimaps, making our method more efficient and applicable to a wider range of scenarios. Additionally, our Click2Trimap is a plug-and-play method that can be seamlessly integrated with trimap-based video matting methods. 

By utilizing Click2Trimap to generate necessary trimaps quickly, the time cost of video matting can be reduced significantly. Quantitative results on video matting datasets are also provided in the section of experiments.

\section{Related Work}
\label{sec:related}
\subsection{Natural Image Matting}

\noindent\textbf{Dense-guided Image Matting.} 

Both before and after the rise of deep learning, trimap-based methods have been the most mainstream and extensively studied methods in the field of natural image matting. Early in this phase, DIM \cite{xu2017deep} built a matting training data pipeline through composite foreground with various background and successfully trained an encoder-decoder model on it. Followup approaches have conducted extensive research on issues including introducing early traditional matting ideas into deep neural networks \cite{tang2019learning,li2020natural}, expansion and augmenting training data \cite{qiao2020attention,li2021deep,li2022bridging,sun2021semantic}, encoder adaptation \cite{park2022matteformer,yao2023vitmatte}, network architectures or modules design \cite{hou2019context,qiao2020attention,lu2019indices,sun2021semantic}, as well as loss function selection and modification \cite{fbaMatting20,lutz2018alphagan,hou2019context}. The trimap-based matting method has the richest targeted research as well as the clearest research route, and also maintains the best performance at present. Trimap-based matting is still the preferred solution for high-precision natural image matting.

Given that manually drawing trimaps is time-consuming, there are also some alternative methods designed to be guided by mask or background. Mask-guided matting methods \cite{yu2021mask,park2023mask} are theoretically easier to extend over a wide range of data domains, but their weakness lies in dealing with targets with large areas of transparency, such as smoke and fire. Matting methods that use the background \cite{sengupta2020background,lin2021real} as guidance have been proposed in recent years, they can efficiently process a series of images with a single background and are used in scenarios such as video conferencing with green screen keying. Due to the specificity of this kind of guidance its use is more limited. 

\noindent\textbf{Sparse-guided Image Matting.} 
Due to the dense guidance information often being difficult to obtain in practical applications, researchers are also dedicated to exploring matting methods that can be guided by sparse information, such as text, clicks, scribbles and so on. CLIPMat \cite{li2023referring} uses textual descriptions as guidance. Utilizing the powerful semantic understanding of multi-modality huge model \cite{radford2021learning}, it achieves qualified performance. UIM \cite{yang2022unified} can be guided by diverse guidance, including clicks, boxes and trimaps. ClickMatting \cite{wei2021improved} and DIIM \cite{ding2022deep} focus on methods that can be guided by the most efficient and flexible form of interaction: clicks. MattingAnything \cite{li2023matting} and MatteAnything \cite{yao2023matte} achieve further enhancements by integrating matting models with SAM \cite{Kirillov_2023_ICCV}. These two methods are currently the most powerful interactive matting methods, yet there is still a significant gap compared to trimap-based matting methods \cite{yao2023vitmatte, liu2023rethinking}.

\subsection{Interactive Segmentation} 
Interactive segmentation is a field dedicated to utilizing user interactions to help models understand various scenarios, which finds widespread applications in various domains, including medical image processing \cite{isegformer, exploring}, image and video editing \cite{anydoor,controlvideo}, and data annotation. Researchers are accustomed to distinguishing between different interactive segmentation models based on forms of interactions. Early methods \cite{xu2016deep,maninis2018deep,mahadevan2018iteratively} used clicks to segment targets. In addition, there are some works devoted to extending the forms of interactions, such as using boxes \cite{rother2004grabcut,xu2017deep,zhang2020interactive} or scribbles \cite{bai2014error,grady2006random,han2022slim} to segment objects. Among all forms of interactions, click-based interactive segmentation models \cite{liu2023simpleclick,sofiiuk2022reviving,liu2022pseudoclick,wei2023focused,chen2021conditional,sofiiuk2020f,chen2022focalclick} are irreplaceable in terms of flexibility and efficiency and also the most well-studied direction. Notably, SimpleClick \cite{liu2023simpleclick} is the first interactive segmentation model to apply the large-scale backbone, ViT-Huge \cite{dosovitskiy2020image}, which resulted in significant performance improvements. This endows interactive segmentation models the ability to build extremely complex feature spaces and strong zero-shot capabilities. More importantly, it has also inspired the exploration of Vision Foundation Model based on interactive segmentation models, such as SAM \cite{Kirillov_2023_ICCV}. Obviously, interactive segmentation models are playing an increasingly important role in computer vision with the pursuit of large visual models.

\section{Method}
\label{sec:Method}

\begin{figure}[t]
    \vspace{-1em}
    \centering
    \includegraphics[width=0.99\textwidth]{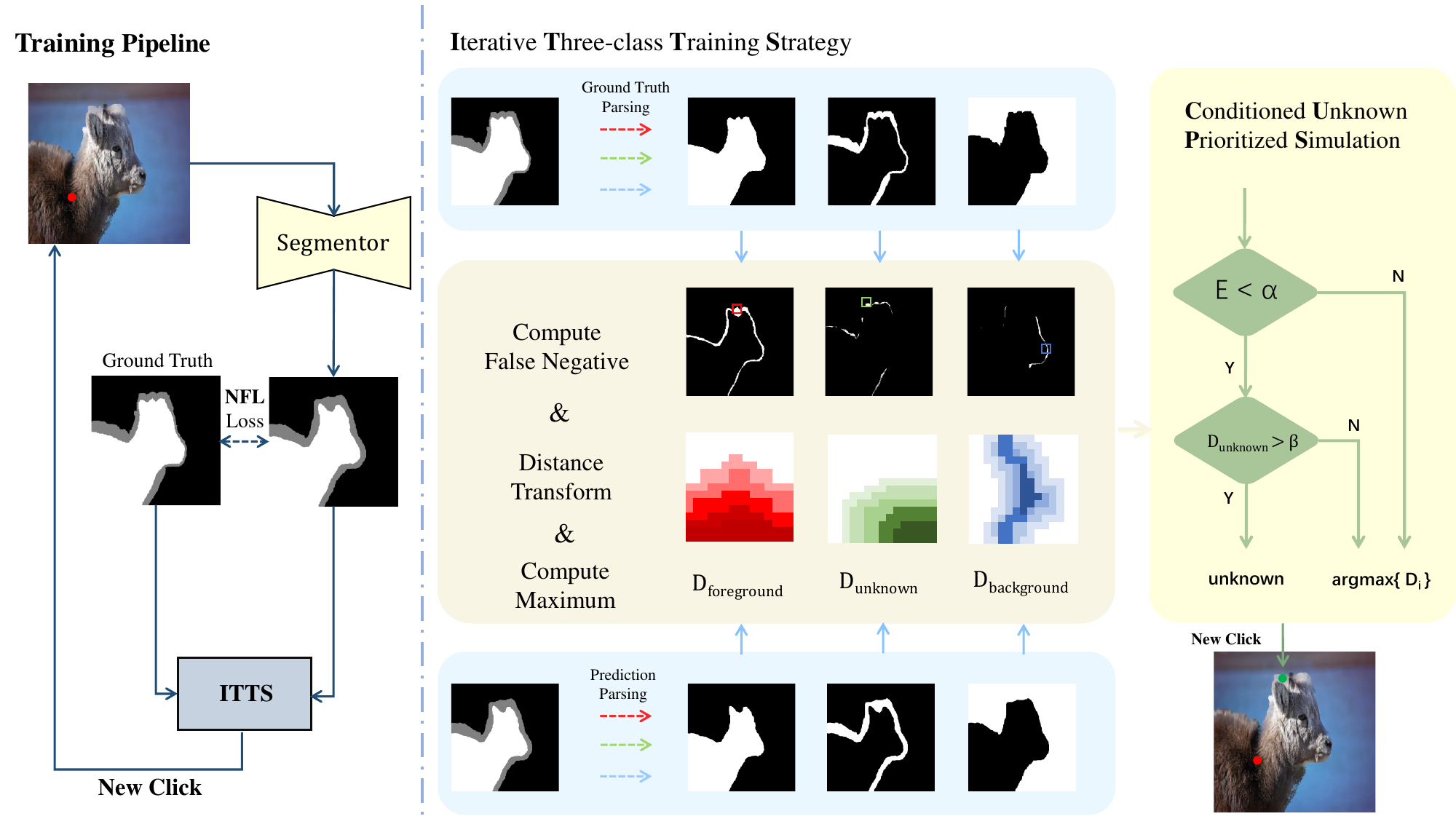}
    \vspace{-0.6em}
    \caption{Left part represents the iterative training loop of Click2Trimap. The role of ITTS, described in right part, is to provide continuous clicks during training. We represent errors by computing false negative error maps and performing distance transform for them. Click2Trimap uses CUPS to decide the class of next click, as formulated in the flowchart.}
    \label{fig:pipeline}
    \vspace{-0.5em}
\end{figure}

The overview of the proposed Click2Trimap approach is shown in \cref{fig:pipeline}. We introduce an Integrative Three-class Training Strategy (ITTS) to facilitate model training for interactive trimap prediction. To underscore the significance of the \textit{unknown} region in the trimap, we additionally propose Conditioned Unknown Prioritized Simulation (CUPS) to prioritize the simulation of clicks for the \textit{unknown} region.

\subsection{Iterative Three-class Training Strategy}
\label{sec:ITTS}

Our method aims to predict high-quality trimaps with minimal user clicks, aligning with applications akin to interactive segmentation models, \eg, SAM \cite{Kirillov_2023_ICCV}. For simplicity, we model the trimap prediction task as a three-class interactive image segmentation task, utilizing trimaps in datasets for model training. Having established this, the next crucial step is devising an effective training strategy, given its pivotal role in the realm of interactive models.

To simulate scenarios where interactive models receive both image data and interactions from real users, prior works have proposed training strategies to mimic real user participation during the training process. An early, simple, and direct training strategy \cite{xu2016deep} involves the dataloader sampling several foreground clicks on foreground regions and an additional set of background clicks on background regions at a given time, based on the ground truth. The model encodes these clicks, sends them to the network along with the image to obtain the prediction mask, and the optimizer updates the model after calculating the error between prediction and ground truth.

However, such a training strategy is not aligned with practical applications~\cite{mahadevan2018iteratively}. In reality, users prefer providing clicks sequentially and may stop once satisfied with the current prediction, deviating from the batched click approach. To address this, ITIS~\cite{mahadevan2018iteratively} introduces the iterative training strategy. Instead of updating the model's parameters after each propagation, a simulation function generates an additional click based on error maps obtained by comparing the current prediction with the ground truth. This click is then sent to the model along with existing clicks to generate a new prediction. The model updates its parameters after several iterations of this process. The iterative training strategy has become a central approach in interactive segmentation, significantly enhancing performance.

Applying this training strategy to the task of trimap prediction is not straightforward. In a binary classification task, which is essentially what interactive segmentation is, false positive and false negative error maps suffice to represent all errors. The simulation function can determine the class of the next click by comparing which error map has the larger error. However, for trimap prediction, a three-by-three confusion matrix is needed to represent prediction errors, resulting in six different error maps containing information on correct and misclassified classes for each pixel. This complexity in error maps is not conducive to the simplicity of simulation rules. More importantly, modeling errors in this way does not align with the logic of real users during interaction. To address these issues, we propose the Iterative Three-class Training Strategy (ITTS) to leverage the concept of iterative training.

Firstly, let's reconsider this problem from the perspective of a real user. In real interaction scenarios, users do not need to know which class the misclassified region has been assigned to; they only need to know what class it should have been to provide the corresponding correct click. In other words, the information provided by false positive error maps is unnecessary for users, and users actually interact based on information provided by false negative error maps. Therefore,
as shown in \cref{fig:pipeline}, we denote the predicted trimap as $P_{i}\in \mathbb{R}^{W\times H}$ and ground truth trimap as $G_{i}\in \mathbb{R}^{W\times H}$, where $i\in\{$\textit{foreground}, \textit{background}, \textit{unknown}$\}$. We then compute three false negative maps using the equation:
\begin{equation}
    FN_{i} = (\lnot P_{i}) \land G_{i},
\end{equation}
where $\lnot$ and $\land$ represent logical NOT operation and logical AND operation, respectively. $FN_{i}\in \mathbb{R}^{W\times H}$. Next, we apply the Distance Transform (dist) to each $FN_{i}$ and record the maximum value from every resulting distance map as $D_{i}$:
\begin{equation}
    \begin{split}
    D_{i} &= \mathrm{Max}(\mathrm{dist}(FN_{i})).
    \end{split}
    \label{3} 
\end{equation}
So we use $\left\{{D}_i\right\}$ to represent the error size of the three classes. By comparing $D_{i}$, we can roughly assess which class has the largest error at this iteration. Consequently, the class of the next click can be described as $\mathrm{argmax}(\left\{{D}_i\right\})$. Finally, we simply need to sample a click on the $FN{i}$ of the corresponding class as the next click to be provided to the model. At this point, one simulation of a real user's interaction is complete. This design not only aligns with the patterns of real users but also simplifies code complexity by discarding three false positive error maps. ITTS allows us to introduce the iterative training strategy from interactive segmentation into the task of trimap prediction. The experiments demonstrate that the trimap predictor trained using this strategy, when combined with trimap-based matting models, significantly outperforms all existing click-based matting methods.

\subsection{Conditioned Unknown Prioritized Simulation}
\label{sec:Conditioned Unknown Prioritized Simulation}

While ITTS is powerful, modeling trimap prediction as a normal three-class interactive segmentation task has limitations due to two distinctive characteristics of trimap prediction. The first consideration stems from a shared insight among researchers working on matting. If the \textit{unknown} class in the trimaps used to guide the matting model covers regions that should be labeled as \textit{foreground} or \textit{background}, it tends to have minimal impact on the final alpha matte, as long as this error is not too large. Conversely, if \textit{foreground} and \textit{background} cover areas that should be labeled as \textit{unknown}, it significantly confuses the matting model. Therefore, the \textit{unknown} class holds a unique role among the three classes, suggesting that we can tolerate a drop in precision for \textit{unknown} but strive to ensure its recall. Hence, considering the unequal importance of the three classes, we introduce Conditioned Unknown Prioritized Simulation (CUPS) to replace the simulation function based on $\mathrm{argmax}$ in ITTS.

CUPS decides the class of next click by evaluating the error level of current prediction trimap at each iteration. Given that we have defined the error size of each class as $D_{i}$ in last section, we directly take $D_{max} = Max(\left\{{D}_i\right\})$ as current largest error size. Meanwhile, we get the region of the target by computing the union of \textit{foreground} and \textit{unknown}. After taking the same operations in Eq. \ref{3}, we define the size of target $D_{t}$ as:
\begin{equation}
    \begin{split}
    D_{t} = \mathrm{Max}(\mathrm{dist}(G_{f} \lor G_{u})).
    \end{split}
\end{equation}

$\lor$ represents the logical OR operation. By computing the ratio of $D_{}max$ to $D_{t}$, we can assess the error level of current prediction trimap:
\begin{equation}
    \begin{split}
    E = \frac{D_{max}}{D_{t}}.
    \end{split}
\end{equation}
When $E$ is less than a preset $\alpha$, CUPS considers the error of current trimap is not significant and will directly provide an \textit{unknown} click for next iteration no matter which class has the largest error. The fundamental purpose of this design is to give \textit{unknown} a prioritization to help all \textit{unknown} regions are recalled. We determine the value of $\alpha$ by complete a parameter search, with more details in the section of experiments.

But this prioritization is not permanent, which is because of the second characteristic 
of trimap prediction task. The ground truth we use are not like the ground truth mask in tasks of segmentation that is absolutely unique and correct. For the same image there are countless trimaps that can be called correct. So when we consider the quality of \textit{unknown} in the predicted trimap is high enough, here we use whether $D_{u}$ is less than the preset $\beta$ to determine this, we remove \textit{unknown}'s priorities. So in CUPS, the class of next click can be formulated as:
\begin{equation}
    \begin{cases}
    \mathrm{unknown}, & E < {\alpha}, D_{\mathrm{unknown}} > {\beta}\\
    \mathrm{argmax}(\left\{{D}_i\right\}), & \mathrm{else}
    \end{cases}
    \label{eq:5}
\end{equation}

\subsection{Training Objectives}
Although we argue the differences between trimap prediction and normal segmentation tasks, the loss functions used in segmentation tasks are more suitable for our task in terms of the forms of output. Moreover, we resize the input resolution to $448 \times 448$ in order to save computation, so the detail-oriented loss functions in matting field may not help our training. We choose Normalized Focal Loss \cite{sofiiuk2022reviving} as our loss function, which is widely used in interactive segmentation methods:
\begin{equation}
\mathrm{NFL}(i, j)=-\frac{1}{\sum_{i, j}\left(1-p_{i, j}\right)^\gamma}\left(1-p_{i, j}\right)^\gamma \log p_{i, j}
\end{equation}
where $p_{i, j}$ denotes the confidence at $(i,j)$ of the prediction $\mathbb{R}^{W\times H\times3}$. Normalized Focal Loss significantly alleviates the problem of fading gradients as predictions become more and more accurate.

\section{Experiments}
\label{sec:exp-sec}

\subsection{Datasets and Metrics}
\label{sec:quants}
\noindent\textbf{Datasets.} We selected four image matting datasets to evaluate our model's performance on natural image matting. Composition-1K \cite{xu2017deep} and AIM-500 \cite{li2021deep} are natural image matting datasets that include challenging and diverse objects such as portraits, glass, plants and plastic. AM-2K \cite{li2022bridging} and P3M-500 \cite{li2021privacy} mainly consists of animals and human portraits. Notably, AIM-500 \cite{li2021deep}, AM-2K \cite{li2022bridging} and P3M-500 \cite{li2021privacy} are all real-world datasets. We choose them to examine the performance of our method in real-world scenarios, which is more in line with practical applications.

\noindent\textbf{Metrics.} For comparison with other interactive matting methods, we evaluate Click2Trimap's performance by evaluating the quality of the alpha matte inferred with trimaps predicted by our method. As mentioned in \cite{qiao2020attention, yao2023matte, yao2023vitmatte, seong2022one}, it's inappropriate to compute metrics only on the $unknown$ region for matting methods which are not guided by trimaps. So we compute all metrics on the whole images like \cite{qiao2020attention, yao2023vitmatte, yu2021mask}. Following the preference in the field of interactive segmentation, we evaluate our model in a manner more consistent with practical applications. Using the corresponding simulation function, we iteratively provide up to ten clicks per image, recording the optimal results achieved during this process—representative of the outcomes users attain when they cease clicking.

\subsection{Implementation Details}
\label{sec:Implement}

We choose ViT \cite{dosovitskiy2020image} initialized with the MAE pretrained weights \cite{he2022masked} as our backbone. The main structure remains similar to SimpleClick \cite{liu2023simpleclick}, but with some necessary adjustments to the fusion of interaction information and channels of output. We train our model on combined dataset for 25 epochs and optimize it with Adam. Initial learning rate is set $5\times10^{-5}$ and decreases to $5\times10^{-6}$ on 20th epoch. We set batchsize to 32 and finished experiments on 4 A100 GPUs. $\gamma$ = 2 in NFL. $\alpha$ and $\beta$ of ITTS is set to 0.1 and 2.

\begin{table}[t]
\small
\caption{Comparison with other interactive matting methods on Composition-1K, AIM-500, AM-2K and P3M-500-NP. Upper part represents trimap-based matting methods and lower part represents trimap-free matting methods.}
\centering 
\begin{tabular}{l *{8}{C{1.0cm}}}

\bottomrule
\multirow{2}{*}{Method} & \multicolumn{2}{c}{Composition-1K} & \multicolumn{2}{c}{AIM-500}  & \multicolumn{2}{c}{AM-2K}  & \multicolumn{2}{c}{P3M-500-NP} \\
\cmidrule{2-9} 
& MSE$\downarrow$ & SAD$\downarrow$ & MSE$\downarrow$ & SAD$\downarrow$ & MSE$\downarrow$ & SAD$\downarrow$ & MSE$\downarrow$ & SAD$\downarrow$ \\
\midrule
\vspace{3pt}{\scriptsize \textcolor{gray}{manually draw trimaps}}
\\
GCA \cite{li2020natural} & 3.35& 35.26 & 12.08 & 34.93 & 1.84 & 8.93 & 1.62 & 7.90 \\
MatteFormer \cite{park2022matteformer} & 1.3 & 23.8 & 8.74 & 26.87 & 1.77 & 8.69 & 1.51 & 7.81 \\
ViT-Matte \cite{yao2023vitmatte} & 1.10 & 20.4 & 3.8 & 17.21 & 1.17 & 7.94 & 1.24 & 7.26 \\
\midrule
\vspace{3pt}{\scriptsize \textcolor{gray}{trimap-free interactions}}
\\
UIMatting \cite{yang2022unified} & 6.0 & 49.9 & - & - & - & - & - & -  \\
MGMatting \cite{yu2021mask} & 2.49 & 33.86 & 57.94 & 134.32 & 76.16 & 169.71 & 36.55 & 91.13\\
MattingAnything \cite{li2023matting} & 26.73 & 124.22 & 14.57 & 43.06 & 3.5 & 17.3 & 9.2 & 25.82 \\
\midrule
\vspace{3pt}{\scriptsize \textcolor{gray}{na\"ive trimaps generation}}
\\
MatteAnything \cite{yao2023matte} & 2.6 & 28.3 & 9.36 & 27.83 & 3.29 & 11.92 & 2.80 & 10.68 \\
\midrule
\vspace{3pt}{\scriptsize \textcolor{gray}{learning trimaps via clicks}}
\\
Click2Trimap (ours) & 1.07 & 22.83 & 4.67 & 20.52 & 1.91 & 12.06 & 1.61 & 10.57 \\
\toprule
\end{tabular}
\label{tab:bigtable}
\end{table}

\subsection{Quantitative Results and Computation}
\label{sec:synthetic results}

After combining with arbitrary trimap-based matting model, Click2Trimap can be used as an interactive matting method based on clicks. We provide trimaps predicted by Click2Trimap to ViT-Matte \cite{yao2023vitmatte} to acquire final alpha mattes here. As demonstrated in Table \ref{tab:bigtable}, Our method significantly outperforms all existing trimap-free matting methods \cite{yang2022unified,yao2023matte,wei2021improved,li2023matting} across synthetic and real-world datasets. 

\begin{figure}[t!]
  \centering
  \begin{minipage}{0.48\textwidth}
    \centering
    \includegraphics[width=\linewidth]{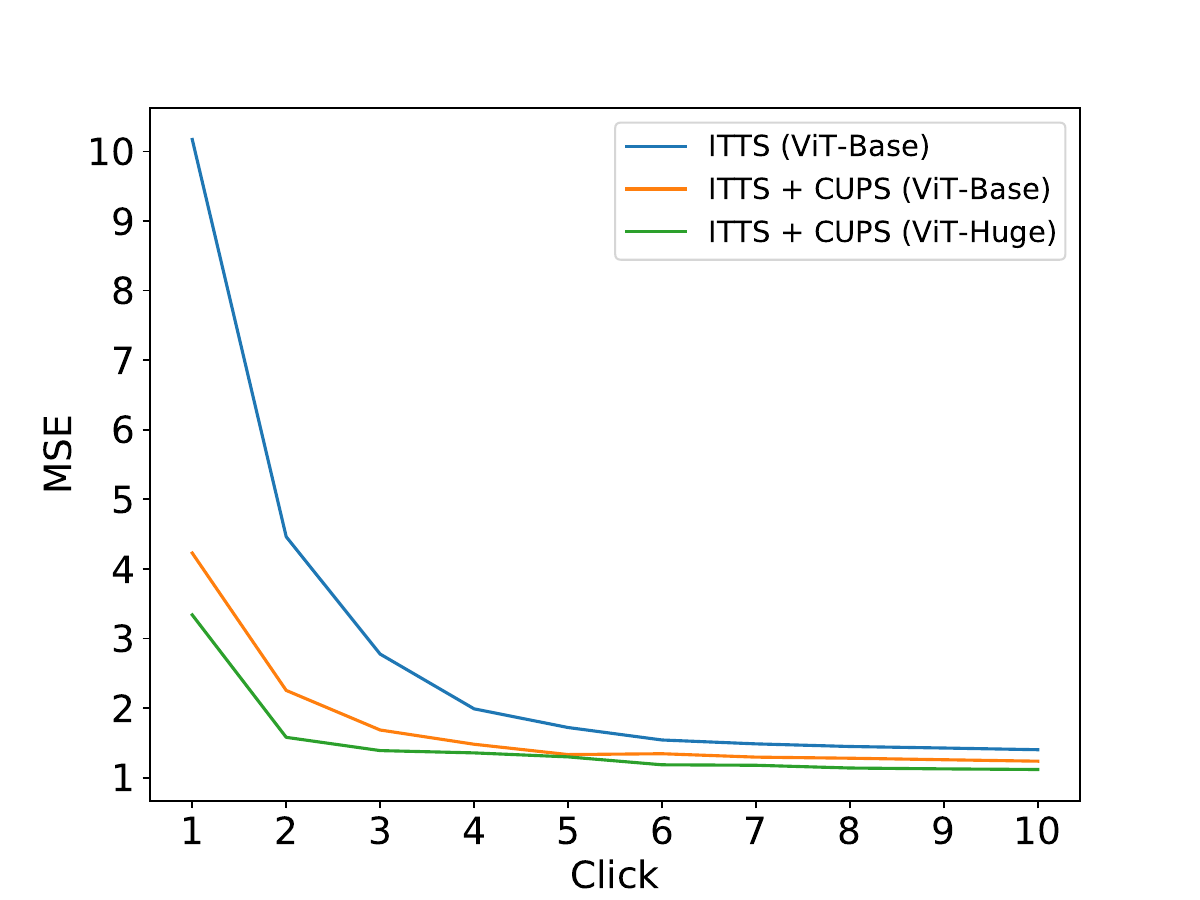}
\caption{We draw this curve by calculate the MSE of the alpha matte guided by the trimap after each click. This obvious declining trend indicates our method not only performs well in predicting trimaps, but also is capable of correcting trimaps continuously.}
    \label{fig:curve}
  \end{minipage}
  \hfill
  \begin{minipage}{0.48\textwidth}
    \centering
\includegraphics[width=\linewidth]{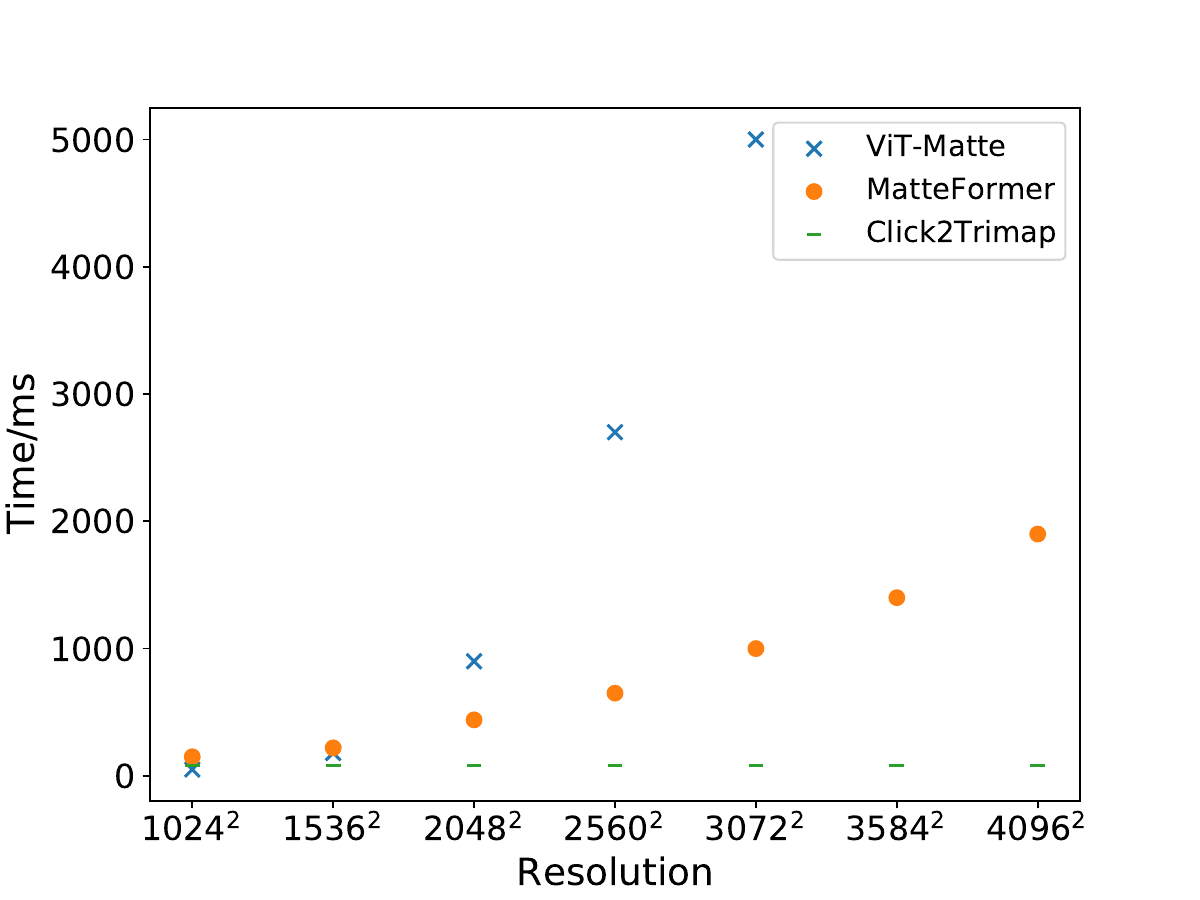}
    \caption{We test the inference time of Click2Trimap (ViT-Huge), ViT-Matte and MatteFormer at different resolutions on one A100 (40G). Even when using a relatively large backbone, Click2Trimap only requires around 80ms to infer a trimap.}
    \label{fig:computation}
  \end{minipage}
\end{figure}

Additionally, we illustrate the correlation between quality of predicted trimaps and the number of clicks in Fig. \ref{fig:curve}. Click2Trimap is proven to achieve accurate results with only minimal clicks and significantly outpaces all existing click-based methods within just 2 clicks.

Since Click2Trimap will directly resize high resolution images to a low resolution like $448\times448$, its computation is notably lower compared to matting models. According to Fig. \ref{fig:computation}, its inference time stands at only 10-20\% of matting models at a resolution of $2000\times2000$ and less than 5\% at a resolution of $4000\times4000$. Therefore, the additional computation introduced by our method is negligible.

\subsection{Ablation and User Studies}
\label{sec:Ablation}
To demonstrate the superiority of ITTS, we provide the result of a version based on two classes for comparison in Table \ref{tab:ablation}, where both $foreground$ and $unknown$ are treated as $foreground$ to adopt the normal iterative training strategy in interactive segmentation. Experiments show that our paradigm almost halves the MSE. And with the addition of CUPS, we further enhance Click2Trimap's performance, achieving high-quality trimap predictions. Additionally, We explore higher performance by increasing the scale of the backbone.

\begin{figure}[t]
  \centering
  \begin{minipage}{0.48\textwidth}
    \centering
    \includegraphics[width=\linewidth]{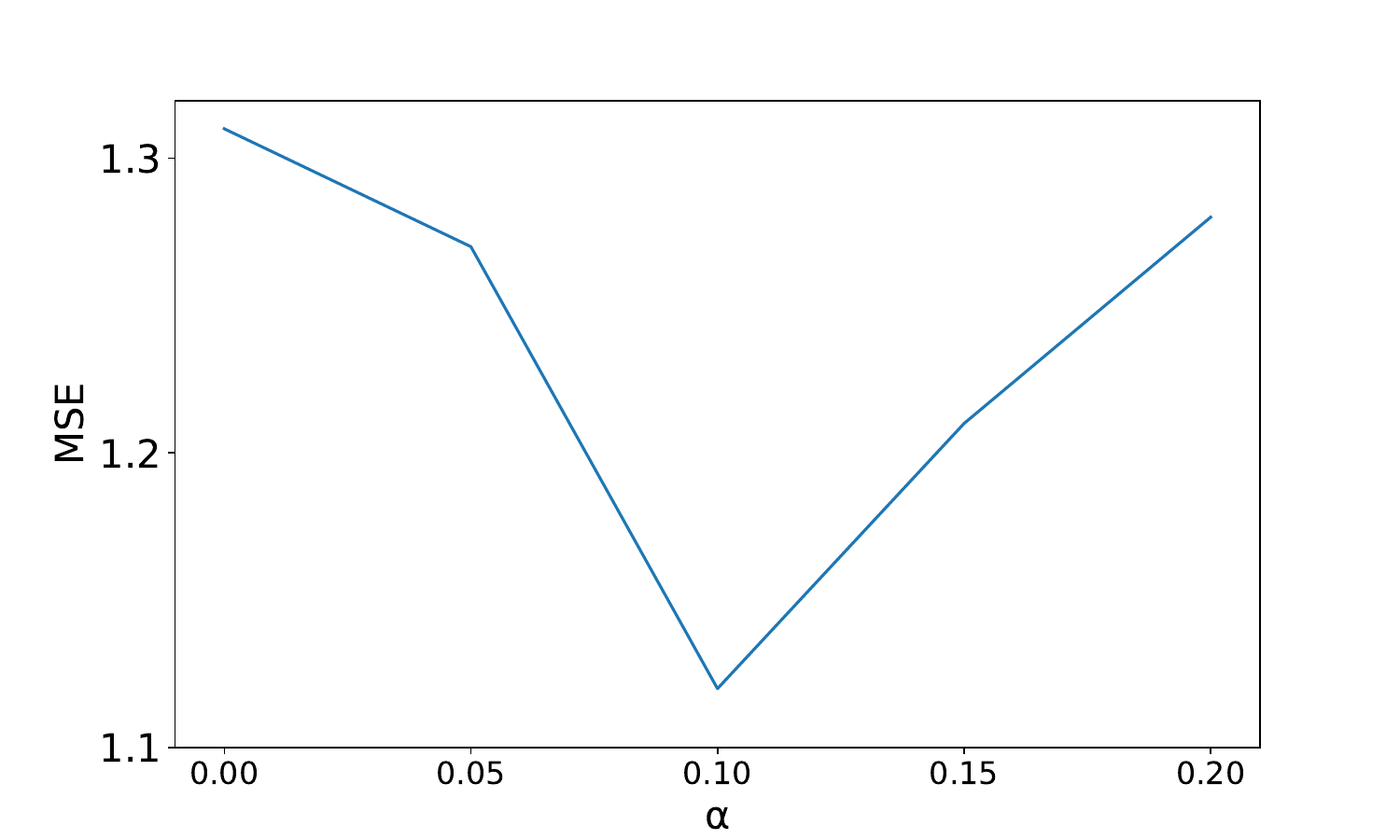}
\caption{Searching process of $\alpha$.}
    \label{fig:searchalpha}
  \end{minipage}
  \begin{minipage}{0.48\textwidth}
    \centering
   \includegraphics[width=\linewidth]{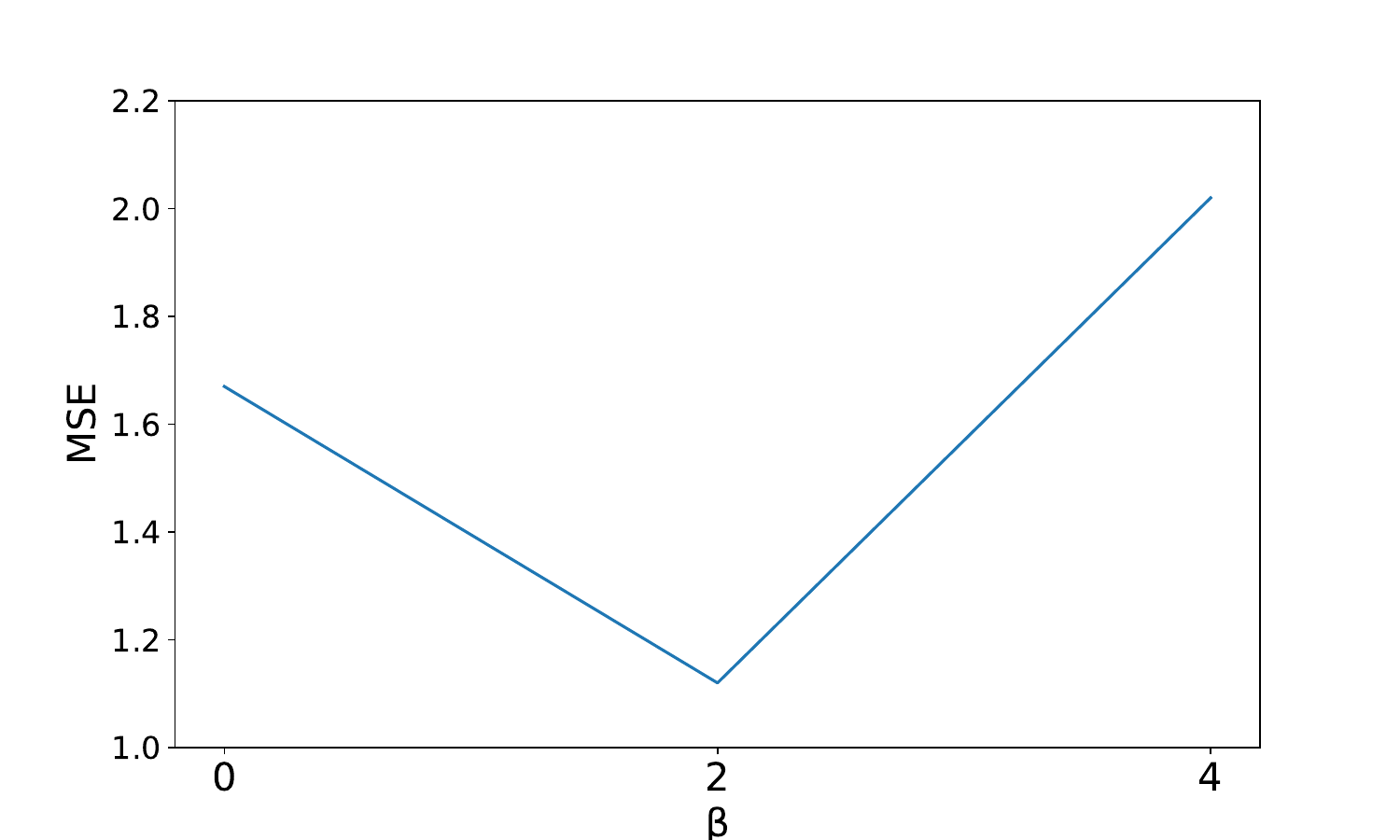}
   \caption{Searching process of $\beta$.}
    \label{fig:searchbeta}
  \end{minipage}
\end{figure}

To determine the values of $\alpha$ and $\beta$, we conducted a parameter search. To save computational resources, we initially estimated the value of $\beta$ to be 2 and then trained our methods five times with five different values of $\alpha$, as shown in \cref{fig:searchalpha}. After determining that $\alpha = 0.1$ is more appropriate, we proceeded to test three different values of $\beta$, as illustrated in \cref{fig:searchbeta}. Results from these experiments indicates that when the approximate prediction error is less than 10\%, prioritizing the $unknown$ becomes more reasonable. It is also evident that after assigning priority to \textit{unknown} regions via the $\alpha$, it's also necessary to revoke this priority through an appropriate $\beta$. Premature or delayed revocation of this priority would lead to a noticeable decline in performance.

We also conduct a user study to demonstrate the efficiency of Click2Trimap in practical usage. We organized 5 users and assigned 50 different cases in test set of Composition-1K \cite{xu2017deep} to each of them. Every user was asked to add clicks iteratively on every image until the user is satisfied with the current trimap predicted by the model. We record the number of clicks and time spent, subsequently calculating the mean values to evaluate the cost of interaction. Table \ref{tab:userstudy} shows that our method helps users to obtain high-quality trimaps with minimal interaction costs, reflected in both the number of clicks and time spent. 5 participants with diverse backgrounds all achieved satisfactory results and the variance of MSE and SAD was only 0.006 and 0.159, respectively, demonstrating Click2Trimap is sufficiently robust across different click patterns and users' behaviors. Among 5 participants, User1 and User3 are matting researchers familiar with the concept of trimap, User4 has a rudimentary understanding of matting, while User2 and User5 had no knowledge of matting or the definition of trimap before the test.

\begin{table}[t!]
\small
\vspace{0em}
\caption{We validate the effectiveness of ITTS, CUPS and backbones of different scales.}
    \vspace{-2em}
  \begin{center}
    \setlength\tabcolsep{4 pt}
    \begin{tabular}{cc|ccc|cc|c|c}
      \bottomrule
      \multicolumn{2}{c|}{Click Class} & \multicolumn{3}{c|}{Training Strategy} & \multicolumn{2}{c|}{Backbone} & MSE$\downarrow$ & SAD$\downarrow$\\
       \hline
       2  & 3  & ITS & ITTS & ITTS + CUPS & Base & Huge & \ & \ \\
        \checkmark & \ & \checkmark \ & \ & \ & \checkmark & \ &2.61 & 28.70 \\
        \ & \checkmark & \ & \checkmark & \ & \checkmark & \ &1.31 & 23.57 \\
        \ & \checkmark & \ & \ & \checkmark & \checkmark & \ &1.12 & 23.50 \\
        \ & \checkmark & \ & \ & \checkmark & \ & \checkmark &1.07 & 22.83  \\
      \toprule
    \end{tabular}
    \label{tab:ablation}
  \end{center}
  \vspace{-2em}
\end{table}

\begin{table}[t]
\small
\caption{The MSE and SAD of final alpha mattes prove that the accuracy of these trimaps is guaranteed.}
   \vspace{-1em}
  \begin{center}
    \setlength\tabcolsep{5 pt}
    \begin{tabular}{c|c|c|c|c|c|c}
      \bottomrule
      \ & User1 & User2 & User3 & User4 & User5 & Average\\ \hline
      Click & 1.7 & 2.08 & 2.04 & 2 & 2.04 & 1.97 \\
      Time & 3.65 & 5.57 & 6.52 & 6.07 & 4.59 & 5.28\\
      MSE & 0.94 & 1.03 & 1.05 & 1.16 & 1.11 & 1.06\\
      SAD & 21.97 & 22.78 & 22.95 & 23.08 & 22.94 & 22.74\\
      \toprule
    \end{tabular}
    \label{tab:userstudy}
  \end{center}
      \vspace{-1.5em}
\end{table}

\begin{table}[t]
\small
\centering
\caption{
Comparison with state-of-the-art video matting methods, including both trimap-based and fully automatic methods. The result of Click2Trimap is obtained by providing predicted trimaps to FTP-VM \cite{huang2023end}. The test set is VM240K \cite{lin2021real} and the benchmark follows FTP-VM \cite{huang2023end}.
}
\label{tab:video}
\setlength\tabcolsep{3 pt}
\begin{tabular}{l|c|ccccc}
\hline
Method & Guidance & MSE$\downarrow$ & MAD$\downarrow$ & Grad$\downarrow$ & dtSSD$\downarrow$ & Conn$\downarrow$\\

\hline
RVM \cite{lin2022robust} & none & 3.02 & 7.71 &  4.27 &  1.83  &  2.64\\
TCVOM \cite{zhang2021attention} & trimap & 1.51 & 2.98 & 3.01  &  1.96 &  1.44 \\
OTVM \cite{seong2022one} & trimap & 0.56 & 4.55 & 1.63 & 1.32 &  0.74  \\
FTP-VM \cite{huang2023end} & trimap & 0.45 & 4.48 &  1.54 &  1.30 &  0.70 \\

\hline
Click2Trimap & click & 0.58 & 4.73 &  1.89 &  1.37 &  0.84 \\
\hline
\end{tabular}
\vspace{-1em}
                    
\end{table}

\subsection{Video Matting}
\label{sec:video}

The portability of Click2Trimap is demonstrated not only by its ability to seamlessly integrate with arbitrary trimap-based matting method but also by its compatibility with trimap-based video matting methods. TCVOM \cite{zhang2021attention} is a trimap-based video matting model that requires users to provide trimaps of several frames to achieve video matting, whereas OTVM \cite{seong2022one} and FTP-VM \cite{huang2023end} only require users to draw a trimap of the first frame to matting the entire video. Obviously, these methods all necessitate that users manually draw at least one trimap to accomplish video matting. However, leveraging Click2Trimap, users can quickly matting the entire video with just a few clicks, which further lowers the cost of video matting. As shown in Table \ref{tab:video}, the quality of video matting within this scheme is also guaranteed.

\subsection{Visualization}
\label{sec:visualization}

Both \cref{fig:fig2} and \cref{fig:compare} are comparisons between Click2Trimap and the most powerful existing trimap-free matting method, MatteAnything \cite{yao2023matte}. In \cref{fig:vis}, we provide plenty of visualization results. The excellent performance on a diverse range of objects including human portraits, animals, plants, plastics, glass, and ornaments, demonstrates the remarkable robustness of Click2Trimap. 

\subsection{Limitation}
\label{sec:Limitation}
\begin{figure}[t!]

\includegraphics[width=0.99\textwidth]{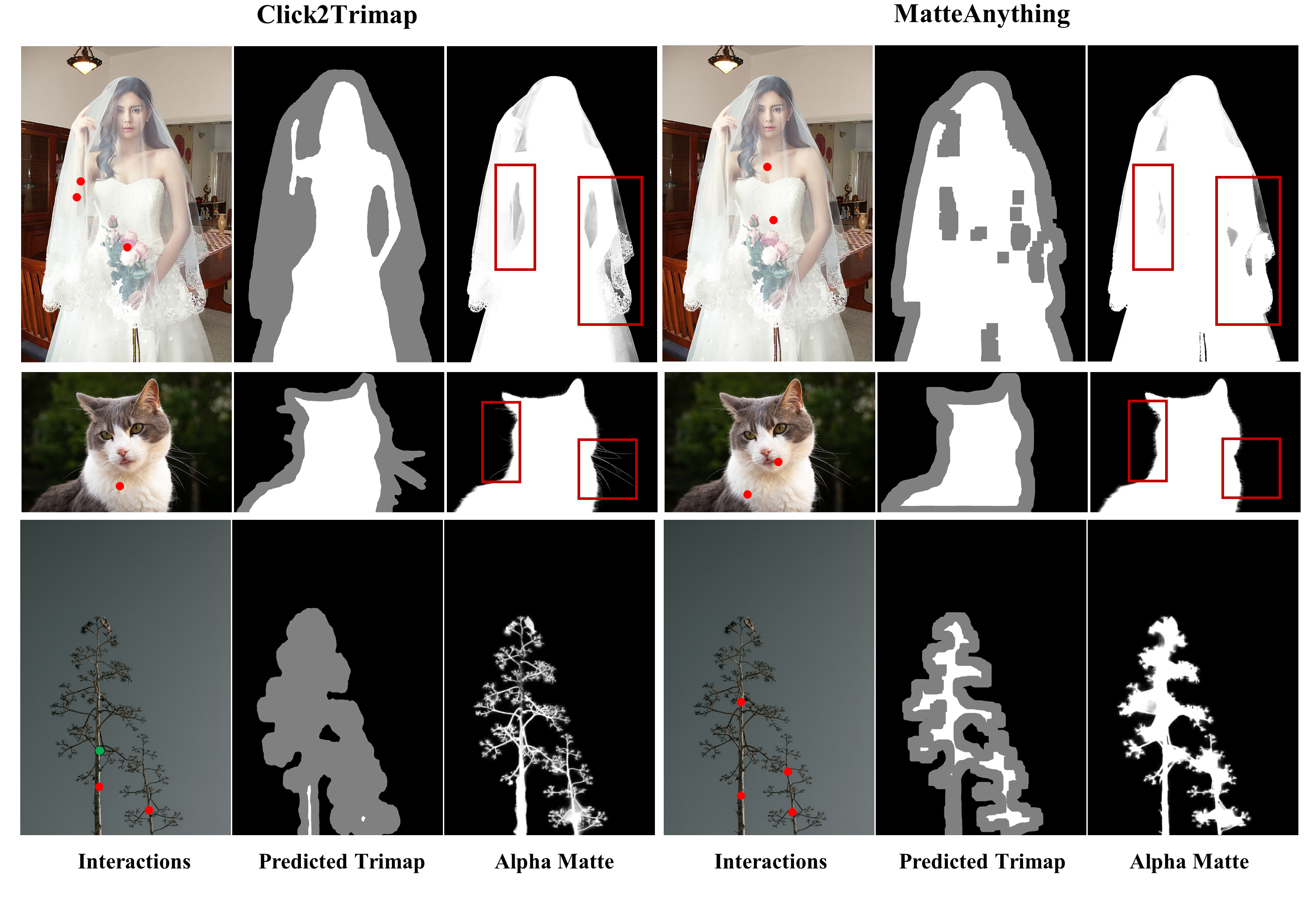}
\caption{Whether facing objects like plastic and wedding dress with large continuous areas of transparency or delicate areas such as a cat's whiskers and tree branches, Click2Trimap can accurately identify them as \textit{unknown}.}
\label{fig:compare}
\vspace{-1em}
\end{figure}

\begin{figure}[t]

\includegraphics[width=0.99\textwidth]{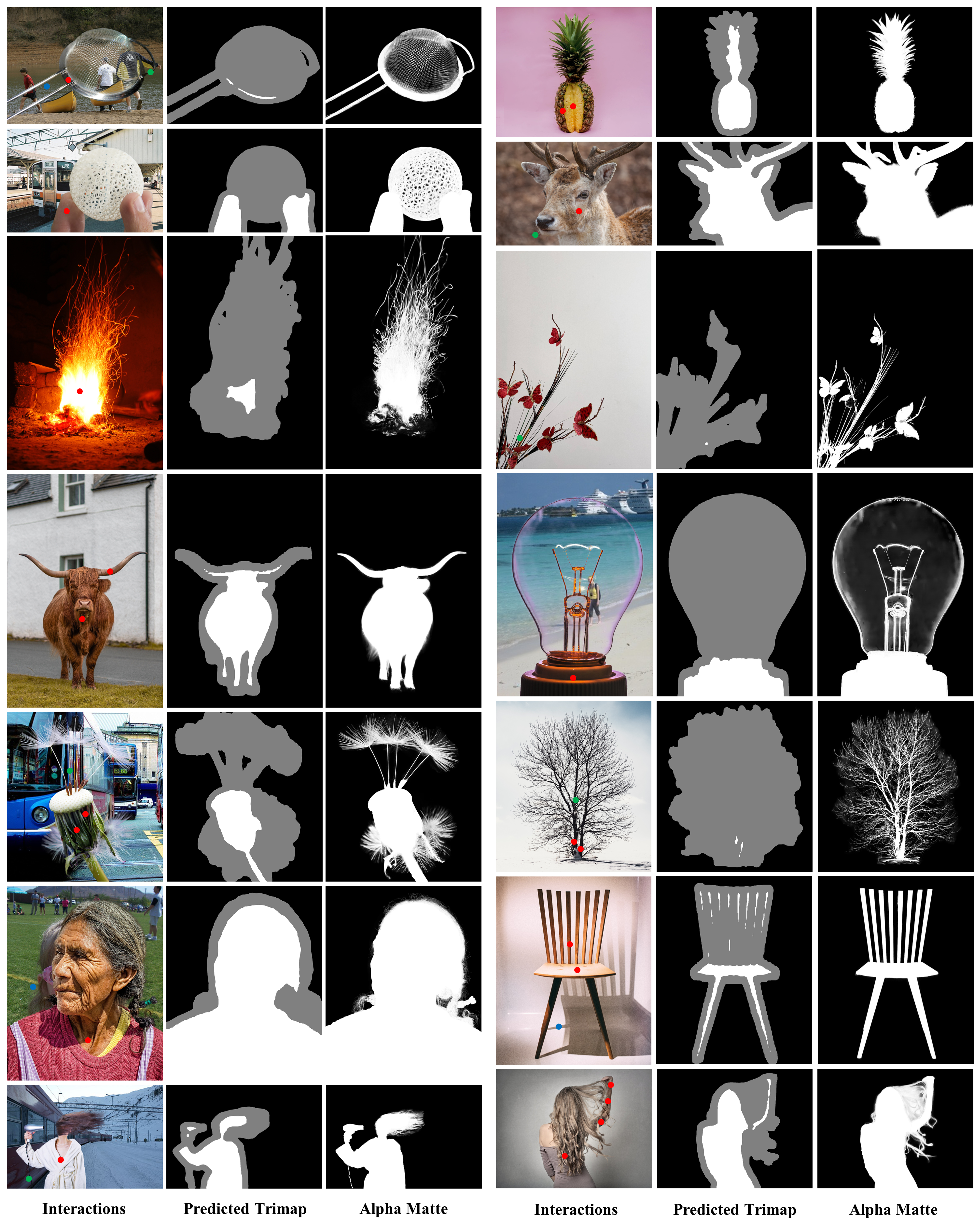}
\caption{We select diverse scenarios to demonstrate the capability of Click2Trimap, including real-world images, synthetic images, and images with relatively complex backgrounds. Furthermore, the dots in the images indicate that Click2Trimap can achieve excellent results with a low interaction cost, which is highly significant for practical applications.}
\label{fig:vis}
\end{figure}

The last case in Click2Trimap \cref{fig:fig2} and \cref{fig:compare} reveal the limitation of Click2Trimap. In these two cases, even after Click2Trimap provide high-quality trimaps, trimap-based matting methods can not predict accurate alpha mattes. And when this happens, Click2Trimap is unable to further optimize the result, even with additional interactions. How to further correct the matting methods under such circumstance is worthy of exploration in the future.

\section{Conclusion}
\label{sec:discussion}

In this paper, we present Click2Trimap, an interactive model capable of predicting high-quality trimaps with just a few clicks. Click2Trimap seamlessly integrates with any trimap-based matting method and can ultimately serve as a click-based matting method. The superiority of Click2Trimap is demonstrated through quantitative and qualitative results across multiple datasets. Our method effectively addresses a common pain point among trimap-based techniques, eliminating the need for users to painstakingly provide manual trimaps for image inference. Consequently, Click2Trimap is considerably more equipment and user-friendly, significantly reducing labor in practical applications.

\clearpage
\newpage

\par\vfill\par

\clearpage  
\bibliographystyle{splncs04}
\bibliography{main}
\end{document}